%% file: main.tex
\documentclass[runningheads]{llncs}
\usepackage{graphicx}
\usepackage{tikz}
\usepackage{comment}
\usepackage{amsmath,amssymb} % define this before the line numbering.
\usepackage{color}

\usepackage{multirow}
\usepackage{algorithm} 
\usepackage{ragged2e}
\usepackage{algpseudocode}
\usepackage{pifont}
\usepackage{xcolor}
\usepackage{xspace}
\usepackage{subcaption}
\usepackage{booktabs}
\definecolor{deemph}{gray}{0.6}
\newcommand{\gc}[1]{\textcolor{deemph}{#1}}
\newcommand{\cmark}{\ding{51}\xspace}%
\newcommand{\xmarkg}{\textcolor{lightgray}{\ding{55}}\xspace}%

\makeatletter
\DeclareRobustCommand\onedot{\futurelet\@let@token\@onedot}
\def\@onedot{\ifx\@let@token.\else.\null\fi\xspace}

\def\eg{\emph{e.g}\onedot} 
\def\ie{\emph{i.e}\onedot} 
 
\def\etc{\emph{etc}\onedot} 
 
\def\etal{\emph{et al}\onedot}
\makeatother

\usepackage[accsupp]{axessibility}  % Improves PDF readability for those with disabilities.

\usepackage[width=122mm,left=12mm,paperwidth=146mm,height=193mm,top=12mm,paperheight=217mm]{geometry}

\usepackage[pagebackref,breaklinks,colorlinks,bookmarks=false]{hyperref}

\begin{document}
% \renewcommand\thelinenumber{\color[rgb]{0.2,0.5,0.8}\normalfont\sffamily\scriptsize\arabic{linenumber}\color[rgb]{0,0,0}}
% \renewcommand\makeLineNumber {\hss\thelinenumber\ \hspace{6mm} \rlap{\hskip\textwidth\ \hspace{6.5mm}\thelinenumber}}
% \linenumbers
\pagestyle{headings}
\mainmatter
\def\ECCVSubNumber{100}  % Insert your submission number here

\title{Training Protocol Matters:\\
Towards Accurate Scene Text Recognition via Training Protocol Searching} % Replace with your title

% INITIAL SUBMISSION 
\begin{comment}
\titlerunning{ECCV-22 submission ID \ECCVSubNumber} 
\authorrunning{ECCV-22 submission ID \ECCVSubNumber} 
\author{Anonymous ECCV submission}
\institute{Paper ID \ECCVSubNumber}
\end{comment}
%******************

% CAMERA READY SUBMISSION
% \begin{comment}
\titlerunning{Towards Accurate STR via Training Protocol Searching}
% If the paper title is too long for the running head, you can set
% an abbreviated paper title here
%

\author{Xiaojie~Chu\inst{1}\and
Yongtao Wang\inst{1}\thanks{indicates the corresponding author.} \and
Chunhua Shen\inst{2} \and
Jingdong Chen\inst{3} \and 
Wei Chu\inst{3}
}
\authorrunning{X. Chu, Y. Wang, C. Shen, J. Chen and W. Chu}
% First names are abbreviated in the running head.
% If there are more than two authors, 'et al.' is used.
%
\institute{Wangxuan Institute of Computer Technology, Peking University\\ \email{chuxiaojie@stu.pku.edu.cn, wyt@pku.edu.cn}\and
 State Key Lab of CAD \& CG, Zhejiang University \\
\and Ant Group\\
% \email{jingdongchen.cjd@antgroup.com, weichu.cw@antgroup.com}
}
% \end{comment}
%******************
\maketitle

\input{chaps/abstract}
\input{chaps/introduction}
\input{chaps/related_work}
\input{chaps/method}

\input{chaps/experiments}

\input{chaps/conclusion}

\clearpage

\bibliographystyle{splncs04}
\bibliography{main}

\clearpage

\input{appendix}
\end{document}

%% file: chaps/abstract.tex
\begin{abstract}
The development of scene text recognition (STR) in the era of deep learning has been mainly focused on novel architectures of STR models. However, training protocol (\textit{i.e.}, settings of the hyper-parameters involved in the training of STR models), which plays an equally important role in successfully training a good STR model, is under-explored for scene text recognition. In this work, we attempt to improve the accuracy of existing STR models by searching for optimal training protocol. Specifically, we develop a training protocol search algorithm, based on a newly designed search space and an efficient search algorithm using evolutionary optimization and proxy tasks. Experimental results show that our searched training protocol can improve the recognition accuracy of mainstream STR models by 2.7\%$\sim$3.9\%. In particular, with the searched training protocol, TRBA-Net achieves 2.1\% higher accuracy than the state-of-the-art STR model (\textit{i.e.}, EFIFSTR), while the inference speed is 2.3$\times$ and 3.7$\times$ faster on CPU and GPU respectively. Extensive experiments are conducted to demonstrate the effectiveness of the proposed method and the generalization ability of the training protocol found by our search method. Code is available at \url{https://github.com/VDIGPKU/STR_TPSearch}
\end{abstract}

%% file: chaps/introduction.tex
\section{Introduction}\label{sec:introduction}
Reading text in scene images is of great significance for a wide range of applications, \eg, multilingual translation, blind navigation, and automatic driving.
Hence, scene text recognition (STR) has attracted much research interest in recent years. Different from document text recognition, scene text recognition is very challenging due to the imperfect imagery conditions in natural images, such as complex backgrounds, low resolution, perspective, and curved distortion, \etc. Recent studies use deep learning models to address these challenges~\cite{chen2021text} and make efforts to design new architectures of STR models for accurate recognition, \eg, rectification module for irregular text images \cite{shi2016robust,shi2018aster,yang2019symmetry}, extraction of visual features \cite{wang2017gated,cheng2017focusing,li2019show}, and sequence modeling of language \cite{litman2020scatter,yu2020towards,fang2021read}. 

However, beyond the model architecture, the training protocol (including data pre-processing operations, the choice of optimizer, and the corresponding hyper-parameter settings, \etc) is also crucial to the final performance of the deep neural network. Baek \etal.~\cite{baek2019wrong} investigate the inconsistencies of training data for STR models, and the performance gaps resulting from these inconsistencies. In this work, we further find that the accuracy of the STR models changes dramatically by simply modifying the pre-processing operations for the training data. For example, 
different data filtering strategies used in~\cite{baek2019wrong} and~\cite{shi2018aster} result in 4.2\% accuracy gap on CRNN~\cite{shi2017end}. 
The above two findings lead us to rethink the importance of the training protocol. Further, two subsequent problems are: \textit{is it feasible and how to find an optimal training protocol for the existing STR models?}

\begin{figure}[t]
\centering
\includegraphics[width=0.7\linewidth]{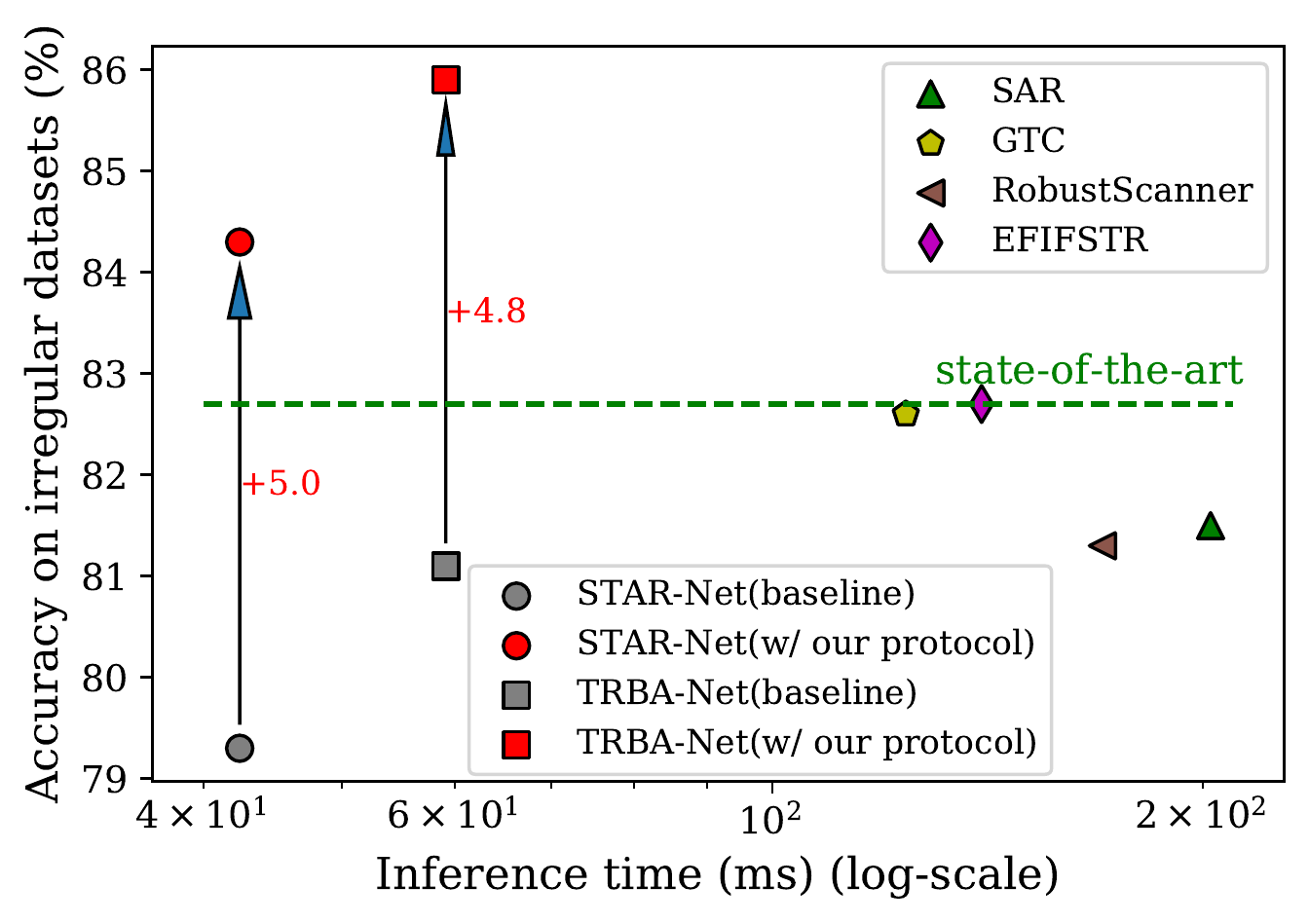}
\vspace{-1em} 
\caption{
Our searched training protocol boosts the accuracy of STAR-Net~\cite{liu2016star} and TRBA-Net~\cite{baek2019wrong}, and achieves new state-of-the-art trade-offs between accuracy and efficiency.
Accuracy are measured on the union of irregular datasets (\ie, IC15, SVT-P and CUTE80) and inference time is measured on Intel i7-6900K CPU. 
}
\label{fig:cpu}
\vspace{-1em} 
\end{figure}

In this work, we make the first attempt to improve the accuracy of scene text recognition by optimizing the training protocol. 
We first summarize the hyper-parameters involved in training an STR model and build a stronger baseline than previous works~\cite{baek2019wrong}. As shown in Table~\ref{table:baseline_abs}, our strong baseline is 5.9\% higher than~\cite{baek2019wrong}. 
To go even further, we propose a training protocol search method to automatically find the optimal training protocol.
Specifically, we first build a search space that contains most popular training protocols used by existing STR methods. Then, we propose an efficient search algorithm to find the optimal training protocol, based on the evolutionary algorithm with a proxy task. 

We conduct quantity and quality experiments to test the effectiveness and generality of the searched protocol. 
As depicted in Figure~\ref{fig:cpu}, the searched optimal training protocol can significantly improve the recognition accuracy of STAR-Net~\cite{liu2016star} and TRBA-Net~\cite{baek2019wrong} by 4.8\% and 5.0\% respectively, \textit{at a low search cost of 8.8 GPU days}. Compared with the state-of-the-art method EFIFSTR~\cite{wang2020exploring}, our trained STAR-Net~\cite{liu2016star} achieves 1.3\% higher accuracy with $3.3\times$ faster inference speed, and TRBA-Net~\cite{baek2019wrong} achieves 3.4\% higher accuracy with $2.4 \times$ faster inference speed.
To sum up, the contributions of this work are as follows:
\begin{itemize}
\item We first systematically investigate the importance of the training protocol for STR task and find that with a proper training protocol, existing simple models (\eg, STAR-Net~\cite{liu2016star}) are able to achieve higher accuracy than recently proposed ones (\eg, EFIFSTR~\cite{wang2020exploring}).

\item We propose a novel and efficient search algorithm to automatically find the optimal protocol for training STR models, at a low search cost of 8.8 GPU days.
\item 
Our searched training protocol shows a good generalization capability over mainstream STR models and can significantly improve their recognition accuracy by 2.7\% $\sim$ 3.9\%.

\item State-of-the-art results (recognition accuracy, as well as the trade-off between recognition accuracy and inference speed) are achieved, by simply applying the searched training protocol to the existing STR models.
\end{itemize}

%% file: chaps/related_work.tex
\section{Related Work}
\subsection{Architecture Design for Scene Text Recognition}
Recent STR methods tackle scene text recognition as an image-based sequence recognition problem and exploit deep learning.
CRNN~\cite{shi2017end} uses VGG~\cite{Simonyan15} and bidirectional LSTM (BiLSTM) to extract visual feature vectors from the image and adopt the CTC loss to predict character sequence. 
To improve the recognition accuracy, authors of \cite{shi2016robust,lee2016recursive} introduce 
an attention mechanism into the sequence prediction module. 
CTC-based and attention-based models are two mainstream STR methods~\cite{baek2019wrong,chen2021text}.

For irregular text recognition, some methods attempt to rectify irregular images into regular ones before recognition.
RARE \cite{shi2016robust} and STAR-Net~\cite{liu2016star} apply a spatial transformer network (STN)~\cite{jaderberg2015spatial} with Thin-Plate-Spline (TPS)~\cite{bookstein1989principal} transformation as an early rectification module to rectify the whole text in image. 
Afterward, some well-designed rectification networks are proposed by introducing  symmetry constraints~\cite{yang2019symmetry}, rectifying individual characters ~\cite{liu2018char,luo2019moran} or iterative rectification pipeline~\cite{zhan2019esir}. Instead of rectifying images, some works recognize the irregular text directly.
For example,
AON~\cite{cheng2018aon}~proposes an arbitrary orientation network to capture character features in four directions. 
SAR~\cite{li2019show} proposes a 2D tailored attention module to tackle the challenge of irregular scene text recognition. 
EFIFSTR~\cite{wang2020exploring} uses attentional glyph generation with trainable font embeddings for learning font-independent features of scene texts.

Baek \etal.~\cite{baek2019wrong} extensively evaluate previously proposed STR modules and discover the best combinations of them, \ie, TPS-ResNet-BiLSTM-Attention (TRBA-Net for short).
To mitigate the limitations of RNN based methods (\eg, time-dependent decoding manner) or improve the robustness of STR models, some recent methods explore novel architectures for the recognizer.
TextScanner~\cite{wan2019textscanner}~proposes a segmentation-based dual-branch framework for scene text recognition. 
SRN~\cite{yu2020towards}~introduces a Transformer-based~\cite{vaswani2017attention} global semantic reasoning module to capture global semantic context through multi-way parallel transmission. 
Fang \etal.~\cite{fang2021read}~propose autonomous, bidirectional, and iterative ABINet for utilizing linguistic knowledge in scene text recognition.
RobustScanner~\cite{yue2020robustscanner} introduces a novel position enhancement branch and a dynamic fusion module to mitigate the errors in contextless scenarios.
Beyond these works, AutoSTR~\cite{zhang2020efficient}~uses neural architecture search technique to find a data-dependent feature extractor for STR. 

Different from these approaches, we aim to improve the performance of the existing STR methods by searching for  optimal training protocols for them. Our approach is orthogonal and complementary to previous works for architecture designing.
In other words, architectural improvements can be used in conjunction with our training strategy to achieve better performance.

\subsection{Training Methods for Scene Text Recognition}
Existing methods for improving the training are often designed for specific STR models. For CTC-based methods, EnEsCTC \cite{liu2018connectionist}~introduces a maximum entropy based regularization and an entropy-based pruning method to prevent over-fitting while Guided training of CTC (GTC~\cite{hu2020gtc}) is designed to learn a better alignment and feature representations. For solving the attention drift phenomenon of attention-based method, FAN~\cite{cheng2017focusing} uses extra pixel-wise supervising information for training a focusing network, while the edit probability loss~\cite{bai2018edit} is proposed to effectively handle the misalignment problem between the training text and the output probability distribution sequence. Moreover, some auxiliary supervision is specifically designed. 
For example, SCATTER~\cite{litman2020scatter} adds intermediate supervisions to train a deeper BiLSTM encoder. SEED~\cite{qiao2020seed} uses the word embedding from a pre-trained language model to supervise the training of a semantics enhanced encoder-decoder framework.
Besides, to solve the low-quality text recognition from the feature-level, PlugNet~\cite{yanplugnet}~develops a pluggable super-resolution unit for auxiliary training to assist the recognition network.

Different from the above-mentioned works which manually design the loss function or auxiliary supervision, we attempt to use the automatic search technique to find the optimal training protocol (\eg, data pre-processing operations, the choice of optimizer, and the corresponding hyper-parameter settings) for the STR task. 

\subsection{Hyper-parameter Optimization (HPO)}
There are many ways to perform hyper-parameter optimization (HPO), including random search~\cite{bergstra2012random}, Bayesian optimization~\cite{falkner2018bohb}, reinforcement learning~\cite{dong2020autohas}, evolutionary algorithms~\cite{friedrichs2005evolutionary,young2015optimizing} and gradient-based methods~\cite{lorraine2020optimizing}. 
Multi-fidelity optimization methods can accelerate HPO optimization by evaluating on a proxy task (\eg, using fewer training epochs~\cite{li2017hyperband} or a subset of data~\cite{klein2017fast}). 
So far, it is unclear if their improvements can be transferred to state-of-the-art deep models on  large-scale datasets. 
Predictor based methods learn a function to map the performance of target model over the hyper-parameters space, \eg, FBNetv3~\cite{dai2021fbnetv3} trains a performance predictor for hyper-parameters, but it is time-consuming since it requires a relatively large number of hyper-parameter evaluations to train the predictor~\cite{dong2020autohas}.

Previous HPO works mostly focus on search algorithm itself for improving searching efficiency. Unlike prior works, this paper focuses more on designing the search space tailored for the STR task. Moreover, extensive experiments show the effectiveness and generality of the searched training protocol, which significantly improves the accuracy of various STR models. These results fill the gap of lacking strong empirical evidence in previous works about HPO.

%% file: chaps/method.tex
\input{tables/HPO_result}

\section{Training Protocol Search for STR} \label{Method}
Our goal is to find a general training protocol that is effective for various STR models, instead of carefully tuning the value of hyper-parameters for a specific model. However, the number of possible training protocols is very large, making exhaustive evaluation impracticable. 
To address this problem, we introduce hyper-parameter optimization technique to efficiently find an optimal training protocol for STR. We follow the general framework of hyper-parameter optimization, which consists of three components: search space (determining the hyper-parameters to be tuned and their domain), search strategy, and performance estimation strategy~\cite{automl}.

Specifically, we first investigate the training protocols used by existing STR models (Sec~\ref{sec:RTP}) and design a training protocol search space for STR tasks (Sec~\ref{sec:OTPS}). 
Then we develop a proxy task to evaluate candidate protocols (Sec~\ref{sec:PES}) and exploit evolutionary algorithm to find the optimal protocol (Sec~\ref{sec:SA}).

\subsection{Review of Training Protocols} \label{sec:RTP}
\input{tables/train}

The overall process of training an STR model is summarized in Algorithm~\ref{algo:train}.
Specifically, the training data (\ie, images and texts in images) used by mainstream methods~\cite{li2019show,hu2020gtc} are  usually sampled from different datasets (\eg, a part of the data is from the synthetic datasets and the rest is from the real-world datasets). 
At each iteration, images and texts are pre-processed as the input and the expected output respectively, and they are used to calculate the gradients. Then the learning rate ($lr$) is adjusted by a predetermined $lr$ schedule and the model parameters are updated by the optimizer based on the optimization schedule and gradients. Finally, the training process terminates after $K$ iterations. 

In this work, we focus more on general training hyper-parameters that can be applied to various STR architectures.
We do not search for the specific training hyper-parameters (\eg, loss) that are designed and only available for specific models.
We systematically investigate and summarize various hyper-parameters involved in training an STR model, which includes data sampling strategy, data pre-processing, optimizer, $lr$, and $lr$ schedule.
Several examples are shown in Table~\ref{table:HPO_result}.

\subsection{Search Space} \label{sec:OTPS}
\input{tables/HPO_searchspace}
The search space defines all candidate training protocols (\ie, hyper-parameters and their domains), and the search algorithm attempts to find the optimal training protocol in the search space. Specifically, as listed in Table~\ref{table:HPO_searchspace}, we consider 8 hyper-parameters (HPs) in total: 

\textbf{HP-1}: $\alpha_{real}$ indicates \textit{the proportion of the data} sampled from real-world datasets, which is used to control the sampling strategy when training with both synthetic datasets and real-world datasets.
Since the number of images in the real-world datasets is much smaller than the synthetic datasets (\ie, 50k \textit{vs.} 16M), sampling randomly from the union of real-world and synthetic datasets may be  sub-optimal for training. 
In this work, we go further to search for the proportion of the training data sampled from real-world datasets (while the others are sampled from synthetic datasets). 

\textbf{HP-2}: \textit{color format of image}.
Existing data sets have different image formats (\eg, MJSynth~\cite{jaderberg2014synthetic} data set is composed of grayscale images, while the  SynthText~\cite{gupta2016synthetic} data set is composed of RGB images).
Some works~\cite{baek2019wrong,yue2020robustscanner} convert input images of all formats into grayscale ones, while others~\cite{li2019show} convert all the input images into RGB format. We will search for the better one of these two kinds of color formats.

\textbf{HP-3}: whether to \textit{keep the aspect ratio} when resizing input images.
Some works~\cite{wang2020decoupled,hu2020gtc} keep the aspect ratio when scaling the image, while others~\cite{cheng2017focusing,baek2019wrong,litman2020scatter} choose to resize the image to a specific size and change the original aspect ratio. Hence, we consider searching for the better one of these two image scaling strategies.

\textbf{HP-4}: whether to perform \textit{data augmentation} on training data. In this work, we consider the data augmentation strategies used in \cite{fang2021read}.

\textbf{HP-5}: \textit{character set} for the decoder.
The final output of the decoder in the model needs to be converted into a character sequence. Under the evaluation that considers only case-insensitive alphanumeric~\cite{shi2018aster,baek2019wrong}, different works choose to use different character sets for training (\eg, case-insensitive training or case-sensitive training with character post-processing). 
For example, \cite{baek2019wrong,litman2020scatter} use a character set only consisting of digits and lowercase letters and treat other characters as background noise.
While~\cite{shi2018aster,yang2019symmetry} add uppercase letters and punctuation characters to the above character set and use post-processing for character filtering during the testing phase.
Consequently, our search space contains four options of character sets. The main differences between them are whether they contain punctuation characters and whether they are case-sensitive.

\textbf{HP-6}: \textit{optimizer}.
We consider two optimizers commonly used in previous STR works: Adadelta~\cite{zeiler2012adadelta} used in~\cite{cheng2017focusing,shi2018aster,baek2019wrong,wang2020decoupled} and Adam~\cite{kingma2014adam} used in~\cite{li2019show,hu2020gtc}.

\textbf{HP-7}: \textit{initial learning rate}. Refer to the implementation details of \cite{cheng2017focusing,shi2018aster,baek2019wrong,wang2020decoupled,li2019show,hu2020gtc} and our experience, we find that the range of appropriate learning rate value for Adadelta and Adam is quite different. Therefore, we design different search domains for each optimizer.

\textbf{HP-8}: \textit{learning rate schedule}. Learning rate adjustment is crucial for training a model~\cite{he2019bag}. 
Exponentially decaying the learning rate is widely used in recent works~\cite{shi2018aster,li2019show,hu2020gtc,wang2020decoupled}. For example, GTC~\cite{hu2020gtc} decays the learning rate by 0.1 for every 30000 iterations (we call it "multi-step decay"). We consider four options of learning rate schedules as listed in Table~\ref{table:HPO_searchspace}.

To sum up, our designed search space contains various hyper-parameters involved in training a STR model, including data-related hyper-parameters~(HP-1 $\sim$ HP-5) and model optimization-related hyper-parameters~(HP-6 and HP-8).
Moreover, there are totally 9,216 candidate training protocols in it.

\subsection{Proxy Task for Performance Estimation} \label{sec:PES}
A training protocol evaluation can be very time-consuming particularly on STR tasks with large-scale training datasets. For example, for a training protocol, TRBA-Net~\cite{baek2019wrong} requires 2.3 GPU days for training while ABINet~\cite{fang2021read} requires more than 130 GPU days. 
In order to reduce the search cost, we design a proxy task to efficiently evaluate the candidate training protocols. In particular, we use a smaller and faster proxy model instead of target models (\eg, STAR-Net~\cite{liu2016star} and TRBA-Net~\cite{baek2019wrong}) for evaluating the effect of training hyper-parameters. 
Similar with the architecture of existing typical STR models, our proxy model uses CNN for visual feature extraction and RNN for sequence modeling. 

Specifically, our proxy task consists of two parts: the proxy model and the proxy training with shorter training iterations and proxy datasets.
In detail, 
1) we use CRNN~\cite{shi2017end} as the proxy model;
2) we reduce the number of iterations required for training to 1/6 of the original~(\ie, from 300k to 50k);
3) we sample 20\% of the origin training set as the proxy training set, 
and use the remaining images from real-world datasets (in the original training set) as the proxy validation set.

\subsection{Evolutionary-Based Search Algorithm} \label{sec:SA}
We exploit the evolutionary algorithm~\cite{guo2020single} to find the optimal training protocol. During each evolution cycle, we use the proxy task (mentioned in Section~\ref{sec:PES}) to quickly evaluate the training protocol candidates. Specifically, we first randomly sample $M_{init}=16$ candidates from our designed search space (mentioned in Section~\ref{sec:OTPS}) and rank their performance with the proxy task.
In the next $\mathcal{T}=10$ iterations, the top-$k$ ($k=4$ in this paper) candidates of the previous generation are selected as parents for producing $M=8$ new candidates by crossover and mutation operation. For crossover, two randomly selected parents are crossed to produce a new one. For mutation, a randomly selected parent mutates its every choice of hyper-parameters with probability 0.2 to produce a new candidate.

%% file: tables/HPO_result.tex
% \begin{table}
% %\small
% \centering % used for centering table
% \caption{Comparison of the training protocols.
% The notations are the same as Table~\ref{table:HPO_searchspace}.
% %\textit{$lr$} denotes learning rate. 
% %\textit{multistep-0.6} means that the learning rate is decayed 0.1 after 60\% of total training iterations. \textit{D}, \textit{L}, \textit{U} and \textit{P} denotes digits, lowercase, uppercase and punctuation respectively.  $\text{warmup iterations}=\text{total iterations} * \alpha_{warmup}$. 'keep aspect ratio' represents that whether or not to keep the aspect ratio when resizing input images
% } % title of Table
% \label{table:HPO_result}
% \begin{tabular}{c|ccc}
% \hline
% {Training protocols} & {Baseline}\cite{baek2019wrong} & {Searched} \\
% \hline
% $\alpha_{real}$ & - & 0.125 \\
% color of image  & grayscale & grayscale\\
% keep aspect ratio & False & False \\
% character set & D+L & D+L\\
% learning rate schedule & const & ms-0.6-0.9 \\
% % $\alpha_{warmup}$ & 0 & 0.05 \\
% optimizer & Adadelta & Adam \\
% initial learning rate & 1 & 5e-4\\
% \hline
% \end{tabular}
% \end{table}

\begin{table*}[t]
% \addtolength{\tabcolsep}{-2.pt}
\centering
\small

\caption{
Comparison of the training protocols (hyper-parameters and their settings). Our baseline is a best hand-designed protocol and can be futher improved by searching.
% Examples of training protocols (hyper-parameters and their settings). 
% We compare the training protocols used in existing STR works with ours.
% $T_{C}$ and $T_{T}$ represents training protocols searched by CRNN and Transformer baseline, respectively.
$\alpha_{real}$ indicates \textit{the proportion of the data} sampled from the real-world datasets when training with both synthetic datasets and real-world datasets.
% and \textit{uniform} means sampling  randomly from the union of real and synthetic datasets. 
\textit{$lr$} denotes learning rate.
%$\text{warmup iterations}=\text{total iterations} * \alpha_{warmup}$. 
%$\text{milestone}=\text{total iterations} * \alpha_{milestone}$. 
\textit{D}, \textit{L}, \textit{U} and \textit{P} denote digits, lowercase, uppercase and punctuation respectively. The recognition accuracy is evaluated on the union of 7 real-world benchmark dataset with CRNN~\cite{shi2017end}.
}
\resizebox{\linewidth}{!}{
\begin{tabular}{l|ccccc|cc}
\toprule
Training             & SAR                & GTC                & RobustScanner               & SRN              & EFIFSTR  & \multicolumn{2}{|c}{Ours}                \\ 
Protocols &\cite{li2019show}  & \cite{hu2020gtc}   & \cite{yue2020robustscanner} & \cite{yu2020towards} & \cite{wang2020exploring}         & Baseline%~\cite{baek2019wrong} 
& \textbf{Searched}              \\ \midrule
$\alpha_{real}$       & 0.0003$^\ast$       & 0.135              & 0.0003$^\ast$            & 0.135                       & 0.0003$^\ast$      & 0.0003$^\ast$              & \textbf{0.125}              \\
Color format of image                  & RGB                & Gray       & RGB                         & RGB                 & RGB      & Gray         & \textbf{Gray}               \\
Keep aspect ratio                   & \cmark             & \cmark             & \cmark                      &  \cmark            & \xmarkg      & \xmarkg      & \xmarkg            \\
Data augmentation                   & \xmarkg            & \xmarkg            & \xmarkg                     & \cmark              & \xmarkg        & \xmarkg     & \cmark             \\
Character   set                        & D+L+U+P            &    D+L    & D+L+U+P                     & D+L                 & D+L+U+P        & D+L         & \textbf{D+L}               \\ %\hline
Optimizer                          & Adam               & Adam               & Adam                        & Adam                & Adam       & Adadelta        & \textbf{Adam}               \\
$lr$                                      & $1 \times 10^{-3}$ & $1 \times 10^{-3}$ & $1 \times 10^{-3}$          & $1 \times 10^{-4}$  & $1 \times 10^{-3}$  & 1  & $\mathbf{5 \times 10^{-4}}$ \\
% $lr$ schedule                    & ms-0.4-0.7         & ms-0.3-0.7     & ms-0.6-0.8                  & constant              & ms-0.3-0.7        & constant      & ms-0.6         \\
$lr$ schedule                    & multi-step         & multi-step     & multi-step                  & constant              & multi-step        & constant      & \textbf{multi-step}         \\
$lr$ decay rate                    & 0.1         & 0.1     & 0.1                  & -              & 0.9        & -      & \textbf{0.1}         \\
$lr$ decay iterations      & [100k, 200k]        & [100k, 200k]     & [180k, 240k]                   & -              & every 10k        & -      & \textbf{[180k]}         \\
\midrule
Accuracy & 82.1\% & 83.6\% & 82.6\% & 82.4\% & 83.4\%  & 84.3\% & \textbf{88.2\%} \\ \bottomrule
\multicolumn{8}{l}{~$^\ast$ indicates using random sampling strategy from the union of synthetic and real datasets.
%(\ie, $\alpha_{real} \approx 5*10^4/(1.6*10^7)$)
} \\
% \multicolumn{8}{l}{~$^\dag$ indicates the default implementation by ours since the original paper did not report its implementation.} \\

\end{tabular}
}

\label{table:HPO_result}
\end{table*}

%% file: tables/train.tex
\begin{algorithm}[t!]
% \caption{Train a STR model.}
\caption{Pseudo-code for training a STR model.}
\begin{algorithmic}
% \State \textbf{Input: } Total training iterations $K$, learning rate $lr$, optimizer, training datasets
% \State \textbf{Output: } Trained model
\For{$\textrm{iteration}=1, \ldots, K$}
\State images, texts $\gets$ sample(datasets)   \Comment{Data sampling}
\State $X, Y\gets$ preprocess(images, texts)
\State $Z\gets$ forward(model, $X$)
\State $\ell \gets$ loss($Z$, $Y$)
\State grad $\gets$ backward($\ell$)
\State $lr$ $\gets$ adjust($lr$)    \Comment{$lr$ schedule}
\State update(model, optimizer, $lr$, grad)
\EndFor
\end{algorithmic}
\label{algo:train}
\end{algorithm}
\vspace{-.5em} 

%% file: tables/HPO_searchspace.tex
\begin{table}[t]
% \small
\centering % used for centering table
\caption{The proposed search space of training protocols. Tuples of three values in parentheses denote the lowest value, highest, and steps. The elements in braces indicate all the choices available during the search. The notations are the same as Table~\ref{table:HPO_result}.  \textit{ms} denotes using multi-step decay $lr$ rate schedule.
When using \textit{ms} \textit{$lr$} schedule, the learning rate is decayed 0.1 once the number of iterations reaches one of the milestones which are given as a list of percentage (e.g., multistep-0.6-0.9 refers to decay \textit{$lr$} at 60\% and 90\% of total training iterations). 
} % title of Table
\label{table:HPO_searchspace}
\setlength{\tabcolsep}{10pt}
\resizebox{\linewidth}{!}{
\begin{tabular}{l l}
\toprule
% \textbf{Training hyper-parameters} & \textbf{Domain} \\
% \textbf{Training} & \multirow{2}{*}{\textbf{Domain}} \\
% \textbf{Hyper-parameters}&\textbf{Domain}\\
\textbf{Hyper-parameters}&\textbf{Domain}\\
\midrule
$\alpha_{real}$ & $(0,0.5,0.0625)$\\
Color of image & \{grayscale, RGB\}\\
%\hline
Keep aspect ratio & \{True, False\}\\
Data augmentation & \{True, False\}\\
%\hline
Character set& \{D+L, D+L+P, D+L+U, D+L+U+P\}\\
% $\alpha_{warmup}$& \{0, 0.01, 0.02, 0.05, 0.1\}\\
%\hline

%\hline
% \hline
Optimizer&\{Adadelta, Adam\}\\
$lr$& \{2, 1.5, 1, 0.5\} for Adadelta and \{10, 5, 2, 1\}*$10^{-4}$ for Adam\\
% $lr$ schedule & \{const, ms-0.6, ms-0.6-0.9, ms-0.3-0.6-0.9\} \\
$lr$ schedule & \{constant, ms-0.6, ms-0.6-0.9, ms-0.3-0.6-0.9\} \\

%\multirow{2}{*}{(optimizer, $lr$)} & {\{(Adadelta, 2),(Adadelta, 1.5),(Adadelta, 1), (Adadelta, 0.5),}\\
% &{(Adam, 5e-3), (Adam, 2e-3), (Adam, 1e-3), (Adam, 5e-4), (Adam, 2e-4), (Adam, 1e-4) \}} \\
\bottomrule
\end{tabular}
}
\vspace{-1mm}
% \label{table:HPO_searchspace}
\end{table}

%% file: chaps/experiments.tex
\section{Experiment and Analysis}
We conduct extensive experiments to evaluate the proposed method on seven widely used scene text recognition benchmarks. 
Moreover, we perform ablation studies to verify the effectiveness of the proposed search algorithm and the searched training protocol.

\subsection{Datasets} \label{sec:datasets}
Following previous works~\cite{li2019show,hu2020gtc,yue2020robustscanner}, we use two public
synthetic datasets, \ie, MJSynth(MJ)~\cite{jaderberg2014synthetic}, SynthText (ST)~\cite{gupta2016synthetic} and the {real-world training images} provided from public datasets (IIIT5K~\cite{mishra2012scene}, SVT~\cite{wang2011end}, IC13~\cite{karatzas2013icdar} , IC15~\cite{karatzas2015icdar}, COCO-Text~\cite{veit2016coco}) for training. 
We test the trained model on 7 benchmarks including four regular scene-text datasets, \ie, ICDAR 2003 (IC03)~\cite{lucas2005icdar}, ICDAR 2013 (IC13)~\cite{karatzas2013icdar}, IIIT 5k-Words (IIIT5k)~\cite{mishra2012scene}, Street View Text (SVT)~\cite{wang2011end}, and three irregular text datasets, \ie, ICDAR 2015 (IC15)~\cite{karatzas2015icdar}, SVT-Perspective (SVT-P)~\cite{quy2013recognizing}, CUTE80 (CUTE)~\cite{risnumawan2014robust}.
Recognition performance is measured by case-insensitive alphanumeric accuracy in lexicon-free mode for all benchmarks~\cite{shi2018aster,baek2019wrong}.
If not specified, we use the union of 7 real-world benchmark datasets~\cite{baek2019wrong} as a unified test set and report the average recognition accuracy. 

\subsection{Implementation Detail}\label{sec:Implementation_detail} 
We follow the implementation of architecture from \cite{baek2019wrong}. 
In particular, following~\cite{hu2020gtc}, we replace the LSTM~\cite{DBLP:journals/neco/HochreiterS97} in the attention-based decoder with GRU~\cite{DBLP:conf/emnlp/ChoMGBBSB14} and fix the hidden dimensions of RNN-based modules as 512.
The baseline and the searched training protocols are shown in Table~\ref{table:HPO_result}. 
All models are trained for 300K iterations with a total batch size of 256.
For inference time assessment, we measure the per-image average clock time (in milliseconds) on Intel i7-6900K CPU and NVIDIA TITAN X (Pascal) GPU respectively. 

\subsection{The Searched Training Protocol}
\subsubsection{Comparison with Existing Training Protocols.}
At first, we build a strong baseline protocol based on~\cite{baek2019wrong}.
We compare our searched protocol with the existing training protocols, including hand-designed ones from existing STR methods and our baseline. 
As shown in Table~\ref{table:baseline_abs}, when trained with our baseline protocol, CRNN~\cite{shi2017end} achieves 5.9\% higher accuracy than previous implementation in~\cite{baek2019wrong}.
As depicted in Table~\ref{table:HPO_result}, our baseline is the best manually designed training protocol. Furthermore, our searched training protocol can make the accuracy of CRNN further increase by 3.9\% over our baseline. These results demonstrate the effectiveness of the training protocol searched by our method.

\subsubsection{Generality to Mainstream STR Models.}
\input{tables/classic_results}
We further test our searched training protocol on various mainstream STR models (\eg, CTC-based and Attention-based models) with different architectures.
More detail about these architectures can be found in~\cite{baek2019wrong}.
Although the proxy model used in our searching phase is CRNN, other models with different architecture can also obtain significant accuracy gains with our searched training protocol.
As shown in Table~\ref{table:classic_results}, compared with the implementation in~\cite{baek2019wrong}, our implementation achieves 8.3\%$\sim$9.8\% higher accuracy for various models.
Under exactly the same training data and architectures, our searched training protocol improves the accuracy of various models by 2.7\%$\sim$3.9\% over our baseline.
These results demonstrate that the searched training protocol has good generality.

\subsubsection{Generality to Transformer-based Models.} 
Different from CRNN~\cite{shi2017end} which use RNN for sequence modeling, some newly designed STR model adopt Trans-former-based module without any RNN structure for high parallelism.
We also conduct experiments on two Transformer-based models, \ie, SRN~\cite{yu2020towards} and ABINet~\cite{fang2021read}, to further test the generality of our searched training protocol. 

Instead of training in two stages as described in the original paper~\cite{yu2020towards}, we directly train SRN~\cite{yu2020towards} in one stage using training settings mentioned in Section~\ref{sec:Implementation_detail}. 
As shown in Table~\ref{table:classic_results}, our searched training protocol brings 2.2\% accuracy gains over SRN.
This result shows that our searched training protocol has good potential to boost the accuracy of the recently proposed transformer-based STR models.

ABINet~\cite{fang2021read} is the latest state-of-the-art transformer-based method which consists of a visual model (VM) and a language model (LM). Specifically, the VM and LM are pre-trained first and then trained jointly. 
We apply our searched training protocol only in the joint training phase. 
We re-implemented the ABINet-SV~\cite{fang2021read} with Pre-LN~\cite{xiong2020layer} and train it with the same data and iterations as mentioned in Section~\ref{sec:datasets}. 
As shown in Table~\ref{table:ABINet}, our implementation achieve a similar accuracy with the original paper (\ie, 90.3\% vs 90.6\%) while reducing training cost by $12.3\times$.

Compared with the original training protocol used by ABINet~\cite{fang2021read}, our searched training protocol brings 1.8\% accuracy gains. This experiment shows that our method can boost the performance of the latest architecture of STR model.
Once again, it demonstrates that our searched training protocol is more effective than the hand-designed ones.

\input{tables/cmp_abinet}
\input{tables/cmp_gtc}

\subsubsection{Compatibility to Existing Training Strategy.} 
We also test the compatibility of our searched training protocol and existing training strategy GTC~\cite{hu2020gtc}. 
As shown in Table~\ref{table:GTC_SRN}, our protocol is superimposed with GTC and improves the accuracy of STAR-Net~\cite{liu2016star} by 2.8\%. This indicates that our work can be treated as a good supplement to the existing works for training STR models.

%%%%%%%%%%TABLE%%%%%%%%%%%%%

 \subsection{Comparison with the State-of-the-art}

We compare the experimental results of our method with the state-of-the-art ones~\cite{li2019show,hu2020gtc,yue2020robustscanner} on 6 widely used scene text recognition benchmarks. The results of IC03 dataset are omitted in this subsection since most methods do not report the results on it.
For fair comparison, we only compare the methods that trained with both synthetic and real-world data since the results of them are usually much better than the ones using only synthetic data~\cite{li2019show,yue2020robustscanner}.

We utilize STAR-Net~\cite{liu2016star} and TRBA-Net~\cite{baek2019wrong} as the representatives of CTC-based and attention-based models, respectively.
As shown in Table~\ref{table:results}, with our searched training protocol, both of them achieve significantly better results in the metrics of inference speed and recognition accuracy.
Specifically, STAR-Net~\cite{liu2016star} has the fastest inference speed in both CPU and GPU. Our trained STAR-Net~\cite{liu2016star} achieves the second-best accuracy, only after our trained TRBA-Net~\cite{baek2019wrong}. Moreover, compared with EFIFSTR~\cite{wang2020exploring}, our STAR-Net~\cite{liu2016star} is $3.3\times$ and $9.9\times$ faster on CPU and GPU respectively, while achieves higher accuracy.
Notably, our trained TRBA-Net~\cite{baek2019wrong} achieves much better accuracy than the state-of-the-art. In particular, the accuracy gains over the EFIFSTR~\cite{wang2020exploring} on the regular and irregular datasets are 1.2\% and 3.4\% respectively, while the inference speed is $2.37\times$ and $3.75\times$ faster on CPU and GPU respectively.
As for each single benchmark, our TRBA-Net also achieves the best results on most of them.

Besides, Figure~\ref{fig:cpu} further compares the accuracy-efficiency performance between our models with the state-of-the-art models on the irregular datasets.
Notably, our trained models (\ie, STAR-Net~\cite{liu2016star} and TRBA-Net~\cite{baek2019wrong}) achieve new state-of-the-art trade-offs between accuracy and efficiency on these challenging datasets.

 \input{tables/results}

\subsection{Ablation Study}
In order to analyze the effectiveness of our optimal training protocol search method and the searched training protocol, we conduct a series of ablation studies as follows. In this section, the recognition accuracy is evaluated on the union of 7 real-world benchmark datasets.

\subsubsection{Our baseline.}
We ablate the important choices our baseline on CRNN~\cite{shi2017end}.
Our baseline training protocol comes from~\cite{baek2019wrong}, but the accuracy of CRNN trained with our baseline is 5.9\% higher than the result reported in~\cite{baek2019wrong} (that is, 84.3\% \textit{vs.} 78.4\%). 
As depicted in Table~\ref{table:baseline_abs}, we first re-implement CRNN and gain 0.4\% accuracy increase by using a batch size of 256 instead of 192 and hidden size of 512 instead of 256. 
Then, different from the training protocol used in~\cite{baek2019wrong} that filters the images with non-alphanumeric characters in SynthText dataset (thus 5.5M out of 7M training images are used), we make full use of all the training data by omitting the special characters from the corresponding labels instead of simply ignoring the corresponding training images. With this improvement, that is, without filtering data with non-alphanumeric characters, the accuracy of CRNN is significantly increased by 4.2\% (\ie, from 78.8\% to 83.0\%). Further, following~\cite{li2019show,hu2020gtc,yue2020robustscanner}, we additionally use the real-world training data from the public benchmark datasets and gain an extra 1.3\% accuracy increase.
\input{tables/baseline_ab}

\subsubsection{The Effect of the Searched Hyper-parameter Settings.}
\input{tables/HPO_ab}
As depicted in Table~\ref{table:HPO_result}, our searched training protocol is quite different from the baseline. Hence we conduct ablation studies on CRNN~\cite{shi2017end} to deeply analyze the effect of the searched hyper-parameter settings which are not the same as the baseline.

First, as shown in Table~\ref{table:HPO_ab}, the proportion of the real-world data $\alpha_{real}$ and data augmentation improve our baseline by 1.1\% and 0.5\% on the accuracy, respectively, while using both of them 2.3\% accuracy improvement is achieved. Second, using the searched optimizer, $lr$ and $lr$ schedule together, the accuracy increases 1\%, while only using the searched $lr$ can only improve the accuracy by 0.5\%. Surprisingly, we see a 0.3\% decrease in accuracy when only using the searched optimizer and $lr$. These results indicate that different hyper-parameters of the optimization process are coupled together thus they should be optimized jointly rather than independently. 

Finally, when we use all the searched hyper-parameter settings, the accuracy increases by 3.9\%, which is higher than the sum of the improvements achieved by using them individually. These results show that our method can effectively find a better training protocol, and it’s necessary to jointly optimize the different hyper-parameters in the training protocol.

\subsubsection{Effectiveness of our proxy task. } 
\input{tables/corr}
 We conduct extra experiments to demonstrate the effectiveness of our proxy task, by testing the ranking correlation scores (\ie, Kendall Tau $\tau$~\cite{kendall1938new}) between it and existing STR methods. Specifically, we first random sample 16 candidate training protocols; and then we get their scores by our proxy task and their fully trained accuracy for CRNN~\cite{shi2017end}, STAR-Net~\cite{liu2016star} and TRBA-Net~\cite{baek2019wrong}, respectively. As shown in Fig.~\ref{fig:corr}, very high ranking correlation scores are achieved, that is, 0.82 for CRNN, 0.75 for STAR-Net, and 0.77 for TRBA-Net. These results show that, our proxy task can effectively predict the performance ranks of the candidate protocols as the fully trained STR models, though with reduced training iterations, sampled datasets, and simple architecture.

\subsubsection{Our Search Method v.s. Random Search.}
\input{tables/rand_baseline}
Following~\cite{li2020random,yu2020hyper},  we test the effectiveness of the proposed search algorithm by comparing it with random search.
Specifically, we randomly sample 32 training protocols to train CRNN~\cite{shi2017end} model with 300K iterations and report the best result of them. 
As shown in Table~\ref{table:rand_baseline}, our search algorithm performs much better than random search, that is, it gains a significantly higher accuracy score while the search cost is much smaller than the random search. 
This experiment demonstrates that our search algorithm is more effective and efficient.

%% file: tables/classic_results.tex
\begin{table}[t]
\renewcommand\arraystretch{1.1}
\small
\centering

\caption{Results of the mainstream STR models when training with our searched training protocol. We also list the results reported in~\cite{baek2019wrong} for reference. 
}
\setlength{\tabcolsep}{2pt}
\resizebox{\linewidth}{!}{
\begin{tabular}{l|ccccccc c}
\toprule
& CRNN & R2AM  & GRCNN  & Rosetta  & RARE  & STAR-Net & TRBA & SRN            \\  
Protocol                & \cite{shi2017end} & \cite{lee2016recursive} & \cite{wang2017gated} & \cite{borisyuk2018rosetta} & \cite{shi2016robust} & \cite{liu2016star} & \cite{baek2019wrong} & \cite{yu2020towards}             \\ \midrule
\cite{baek2019wrong} & 78.4              & 78.5                    & 79.8                 & 80.0                         & 82.0                   & 82.9               & 84.0                   & -                    \\ 
Baseline             & 84.3$^\mathbf{+5.9}$     & 84.6$^\mathbf{+6.1}$           & 85.3$^\mathbf{+5.5}$        & 85.5$^\mathbf{+5.5}$              & 88.3$^\mathbf{+6.3}$        & 88.1$^\mathbf{+5.2}$      & 89.8$^\mathbf{+5.8}$        & 88.5                 \\
Ours                 & 88.2$^\mathbf{+9.8}$     & 87.7$^\mathbf{+9.2}$           & 89.0$^\mathbf{+9.2}$        & 88.4$^\mathbf{+8.4}$              & 91.0$^\mathbf{+9.0}$        & 91.2$^\mathbf{+8.3}$      & 92.7$^\mathbf{+8.7}$        & 90.7                 \\   
\bottomrule       
\end{tabular}
}
\vspace{-1.5mm}
\label{table:classic_results}
\end{table}

%% file: tables/cmp_abinet.tex
\begin{table}[t]
\small 
\centering
\caption{Results of ABINet-SV~\cite{fang2021read} when training with different training protocols. \textit{VM} and \textit{LM} denote the vision model and the language model in ABINet~\cite{fang2021read} respectively. \textit{Joint} represents training \textit{VM} and \textit{LM} jointly. 
The original result of ABINet-SV is reported for reference (\gc{gray}). 
} 

\setlength{\tabcolsep}{3pt}
\begin{tabular}{c|ccc|cccc|c}
\toprule
Training            & \multicolumn{3}{c|}{Training Data} & \multicolumn{4}{c|}{Training time~(GPU days)} & \multirow{2}{*}{Accuracy (\%)} \\
Protocol            & ST+MJ & WikiText~\cite{merity2016pointer}            & Real           & VM        & LM       & Joint      & Total      &                           \\  \midrule
\multirow{2}{*}{\cite{fang2021read}} & \gc{\cmark}   & \gc{\cmark}    &     & \gc{9.3}       & \gc{100}      & \gc{21.3}       & \gc{130.6}      & \gc{90.6}                      \\  
 &  \cmark &  & \cmark         & -         & -        & 10.6       & 10.6       & 90.3                      \\ \midrule
Ours      & \cmark          &          & \cmark         & -         & -        & 10.6       & 10.6       & 92.1                      \\\bottomrule   
\end{tabular}
\vspace{-1.5em} 
\label{table:ABINet}
\end{table}

%% file: tables/cmp_gtc.tex
\begin{table}[t]
\centering
\small 
\setlength{\tabcolsep}{3pt}
\caption{
Results of STAR-Net~\cite{liu2016star} when training with different protocols.
}
\begin{tabular}{c|cccc}
\toprule
Training   Protocol  & Baseline             & Baseline+GTC~\cite{hu2020gtc} & Searched  & Searched+GTC~\cite{hu2020gtc} \\  \midrule
Accuracy~(\%)        & 88.2                 & 88.7 (+0.5)   & 91.2 (+3.1)                & \textbf{91.5 (+3.3)}           \\ \bottomrule
\end{tabular}
\label{table:GTC_SRN}
\vspace{-1.5em} 
\end{table}

%% file: tables/results.tex
\begin{table*}[t]
\centering
\caption{ Results of recognition accuracy on 6 widely used benchmarks (without IC03).
Number of samples in each dataset is shown below the name of dataset. 
}

\small 
\resizebox{\linewidth}{!}{
\begin{tabular}{l|cc|ccc|c|ccc|c|c}
\toprule
\multicolumn{1}{c|}{\multirow{2}{*}{Method}} & \multicolumn{2}{c|}{Time (ms)} & IIIT5K        & SVT           & IC13          & Regular       & IC15          & SVT-P         & CUTE        & Irregular     & Total         \\
\multicolumn{1}{c|}{}                        & CPU            & GPU          & 3000          & 647           & 1015          & 4662          & 2077          & 645           & 288           & 3010          & 7672          \\ \midrule
TextScanner~\cite{wan2019textscanner}$^\ast$      & -              & -            & 95.7          & 92.7          & 94.9          & 95.1          & 83.5$^\dagger$             & 84.8          & 91.6          & -             & -             \\
RobustScanner~\cite{yue2020robustscanner}   & 170.2          & 51.1         & 95.4          & 89.3          & 94.1          & 94.3          & 79.2          & 82.9          & \textbf{92.4} & 81.3          & 89.2          \\
SAR~\cite{li2019show}                       & 202.6          & 31.4         & 95.0          & 91.2          & 94.0          & 94.3          & 78.8          & 86.4          & 89.6          & 81.5          & 89.2          \\
GTC~\cite{hu2020gtc}                        & 124.0          & {13.6}   & 95.5          & 93.2          & 94.3          & 94.9          & 80.4          & 85.5          & {92.0}    & 82.6          & 90.1          \\
EFIFSTR~\cite{wang2020exploring}            & 140.1          & 83.3         & {95.8}    & 91.3          & {95.1}    & 95.0          & 80.9          & 86.0          & 88.5          & 82.7          & 90.2          \\ \midrule
STAR-Net~\cite{liu2016star} +   ours        & \textbf{42.4}  & \textbf{8.4} & 95.7          & {94.1}    & 94.3          & {95.2}    & {82.8}    & {85.1}    & 89.6          & {84.0}    & {90.8}    \\
TRBA-Net~\cite{baek2019wrong} +   ours          & {59.1}     & 22.2         & \textbf{96.6} & \textbf{95.5} & \textbf{95.5} & \textbf{96.2} & \textbf{84.4} & \textbf{89.9} & 90.3          & \textbf{86.1} & \textbf{92.3}
 \\
\bottomrule
\multicolumn{12}{l}{~$^\ast$ indicates using both word-level and character-level annotations for training.} \\
\multicolumn{12}{l}{~$^\dagger$ TextScanner~\cite{wan2019textscanner} use different version of IC15 which contains 1811 images.} \\
\end{tabular}}
\vspace{-2em} 
\label{table:results}
\end{table*}

%% file: tables/baseline_ab.tex
\begin{table}[t]
\centering
\small
\caption{
Ablation study of our baseline training protocol. 
Our baseline is significantly better than training protocol in~\cite{baek2019wrong}.
} 
\setlength{\tabcolsep}{6pt}
\begin{tabular}{lc}
\toprule
Training protocol                              & Accuracy~(\%) \\ \midrule
Training protocol in \cite{baek2019wrong} & 78.4 \\   
Our re-implementation & 78.8$^{+0.4}$  \\ 
~~$+$ data with non-alphanumeric characters & 83.0$^{+4.6}$  \\ 
~~$+$ real-world data                                    & 84.3$^{+5.9}$     \\ \bottomrule
\end{tabular}
\vspace{-2em} 

\label{table:baseline_abs}
\end{table}

%% file: tables/HPO_ab.tex
\begin{table}[t]
\caption{
Ablation study of the searched training protocol. It is necessary to jointly optimize the different hyper-parameters in the training protocol.
}
\setlength{\tabcolsep}{6pt}
\centering
\small
\begin{tabular}{c|l|cc}
\toprule
No. & \multicolumn{1}{c|}{Training protocol}                  & Accuracy & $\Delta$Accuracy \\ \midrule
  & Baseline                                      & 84.3\% & -            \\ \midrule
(1) & $+$  searched $\alpha_{real}$      & 85.4\% & +1.1\%         \\
(2) & $+$  data augmentation                         & 84.8\% & +0.5\%         \\
(3) & $+$  (1) and (2)                              & 86.6\% & +2.3\%         \\ \midrule
(4) & $+$ searched optimizer and $lr$ & 84.0\% & -0.3\%         \\
(5) & $+$ searched $lr$ scheduler      & 84.8\% & +0.5\%         \\
(6) & $+$  (4) and (5)                              & 85.3\% & +1.0\%         \\ \midrule
  & $+$  (1), (2), (4), (5)                              & 88.2\% & +3.9\%        \\
  \bottomrule
\end{tabular}

\vspace{-1em} 
\label{table:HPO_ab}
\end{table}

%% file: tables/corr.tex
\begin{figure}[t]
     \centering
    \vspace{-1mm}
     \begin{subfigure}[b]{0.5\textwidth}
         \centering
         \includegraphics[width=\textwidth]{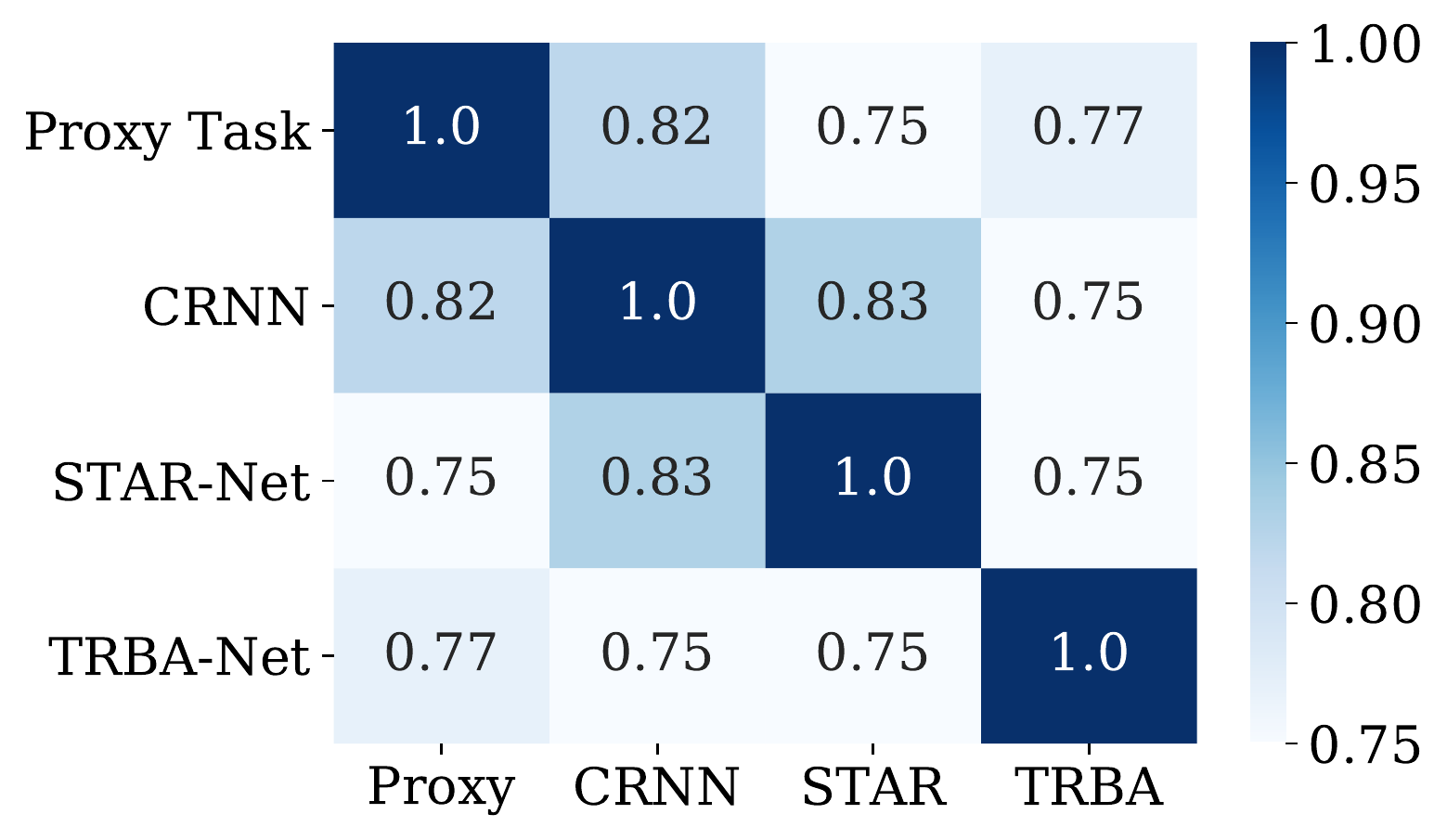}
         \caption{Kendall Tau $\tau$ between different methods.}
         \label{fig:corr_STARNet}
     \end{subfigure}
     \hfill
     \begin{subfigure}[b]{0.4\textwidth}
         \centering
         \includegraphics[width=\textwidth]{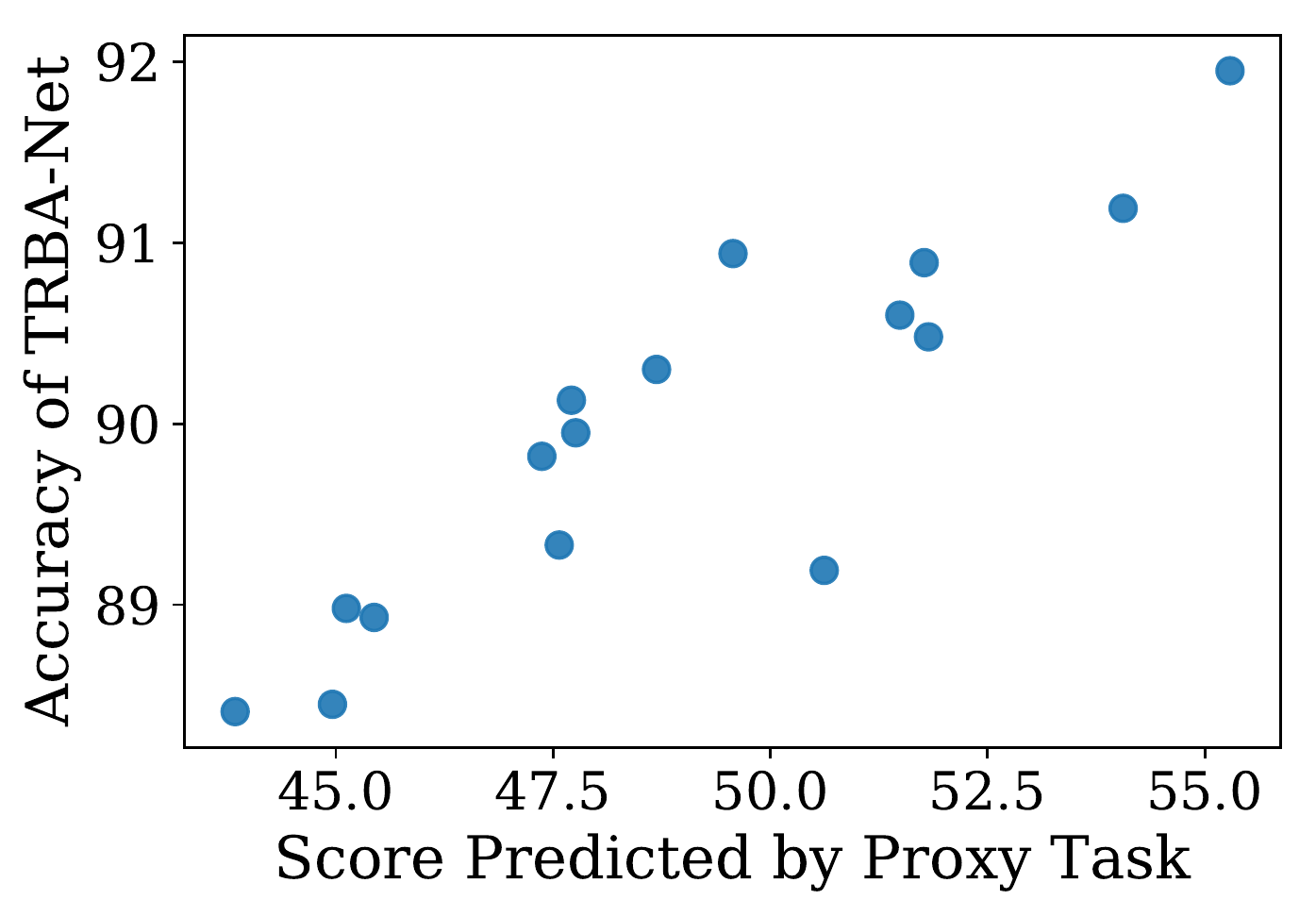}
         \caption{Visualization of correlation between proxy task and TRBA-Net.}
         \label{fig:corr_TRBA}
     \end{subfigure}
     \vspace{-1em} 
        \caption{
        Ranking correlation analysis between our proxy task and existing STR methods.
        }
        \label{fig:corr}
\vspace{-1.5em} 
\end{figure}

%% file: tables/rand_baseline.tex
\begin{table}[t]
\centering
% \small
\caption{Comparisons of the search algorithm. 
} 
\setlength{\tabcolsep}{8pt}
\begin{tabular}{ccc}
\toprule
Search Method & Search Time          & Accuracy             \\
\midrule
Random search w/o proxy task & 19.2 GPU days &  86.3\% \\
{Ours} & {8.8 GPU days} & {88.2\%}$^{+1.9\%}$  \\ 
\bottomrule
\end{tabular}
\vspace{-1em} 

\label{table:rand_baseline}
\end{table}

%% file: chaps/conclusion.tex
\section{Conclusion}
In this paper, we systematically investigate how to improve the recognition accuracy of STR models by optimizing the training protocol for them. Firstly, we summarize the hyper-parameters involved in training a STR model and build a strong baseline than previous works~\cite{baek2019wrong}. Secondly, we build a search space that contains most training protocols used by existing STR methods, and efficiently find the optimal training protocol in it with a newly proposed proxy task and the evolutionary algorithm. Experimental results demonstrate our searched training protocol can boost the recognition accuracy of mainstream STR models with different architectures and can achieve state-of-the-art results on multiple benchmarks. Additional ablation studies further demonstrate the effectiveness of the proposed method for training protocol search, as well as its key components.

%% file: appendix.tex
\appendix
%%%%%%%%% TITLE - PLEASE UPDATE
\title{Appendix}  % **** Enter the paper title here

% \maketitle
% \thispagestyle{empty}
\appendix

%%%%%%%%% BODY TEXT - ENTER YOUR RESPONSE BELOW
\section{Implementation Details}
\subsection{Dataset}

We exploit the following datasets in our experiments, which are widely used in the existing works.

\textbf{MJSynth}~\cite{jaderberg2014synthetic} 
contains 9 million synthetic images generated by rendering words onto natural images with random geometrical transformations. 

\textbf{SynthText}~\cite{gupta2016synthetic} is a synthetic dataset for scene text detection task. We use 7 million cropped word images according to the ground truth word bounding boxes.

\textbf{ICDAR 2003} (IC03)~\cite{lucas2005icdar} 
contains 1,156 images for training and 1,110 images for evaluation.
We do not use its train data because some of them are duplicated in the evaluation dataset of IC13. 
Following~\cite{wang2011end,baek2019wrong}, we discard images that contain non-alphanumeric characters or have less than three characters, which reduces the number of images for evaluation to 867.

\textbf{ICDAR 2013} (IC13)~\cite{karatzas2013icdar} 
contains 848 images for training and 1,095 images for evaluation. 
After filtering the images that contain nonalphanumeric characters, the dataset for evaluation contains 1015 images.

\textbf{IIIT 5k-Words} (IIIT5k)~\cite{mishra2012scene} contains 2,000 images for training and 3,000 images for evaluation, which are collected from Google image search. 

\textbf{Street View Text} (SVT)~\cite{wang2011end} 
is collected from the Google Street View. It consists of 257 images for training and 647 images for evaluation. Many images in SVT are severely corrupted by noise and blur, or with very low resolution.

\textbf{ICDAR 2015} (IC15)~\cite{karatzas2015icdar} 
focuses on incidental scene text, containing a lot of irregular text. 
There are totally 4,468 images for training and 2,077 images for evaluation. 
Previous papers~\cite{cheng2017focusing,bai2018edit,wan2019textscanner} have only used 1,811 images, discarding non-alphanumeric character images and some extremely rotated, perspective-shifted, and curved images for evaluation. However, we use the unfiltered version, which contains 2,077 images, for evaluation by default.

\textbf{SVT-Perspective} (SVT-P)~\cite{quy2013recognizing} 
is collected from the side-view images in Google Street View. Thus, most images are heavily deformed by perspective distortion. The dataset consists of 645 cropped images for testing.

\textbf{CUTE80} (CUTE)~\cite{risnumawan2014robust} 
focuses on curved text recognition. The dataset contains 288 cropped natural word images for testing. 

\textbf{COCO-Text}~\cite{veit2016coco} 
contains 62,351 image blocks cropped from the COCO dataset \cite{lin2014microsoft}. The dataset is partitioned into 42,618 images for training and 19,733 images for validation and testing. 
In this paper, we only use the training set of COCO-Text for training.

\subsection{Evolutionary-Based Search Algorithm}
We exploit the evolutionary algorithm~\cite{guo2020single} to find the optimal training protocol based on our designed search space $\Lambda$ (as mentioned in Section 3.2).
The algorithm is elaborated in Algorithm~\ref{alg:evolution}. 
During each evolution cycle, we use the proxy task to quickly evaluate the training protocol candidates. In this work, we set number of initial candidates $M_{init}=16$, population size $M = 8$, max iterations $\mathcal{T} = 10$, parent population size $k = 4$ and mutation probability $prob=0.2$. We sample 20\% of the origin training set as the proxy training dataset $\mathrm{D}_{train}$, and use the remaining images from real-world datasets (in the origin training set) as the proxy validation dataset $\mathrm{D}_{val}$.

\input{tables/evolution}

\subsection{Label Processing}
The final output of the decoder in the model needs to be converted into a character sequence. 
Under the evaluation that considers only case-insensitive alphanumeric~\cite{shi2018aster,baek2019wrong}, different works choose to use different character sets for training (\eg, case-insensitive training or case-sensitive training with character post-processing). 
Baek \etal.~\cite{baek2019wrong} use a character set only consisting of digits and lowercase letters and filters the images with non-alphanumeric characters in SynthText dataset (thus 5.5M out of 7M training images are used). 
In order to make full use of all the training data, we omit the special characters from the corresponding labels instead of simply ignoring the corresponding training images. 
For example, in Table~\ref{Fig:label_processing}(a), the text in the image contains both uppercase letters and punctuation characters. 
Baek \etal.~\cite{baek2019wrong} will discard this data while we process the label under different character set as shown in Table~\ref{Fig:label_processing}(b).

\subsection{Experiments Details}
\textbf{Training loss.} CTC-based models are trained with connectionist temporal classification loss~\cite{graves2006connectionist} and attention-based (including RNN-based with attention mechanism and Transformer-based) models are trained with cross-entropy loss. 

\textbf{Transformer-based models.} 
In the original paper of SRN~\cite{yu2020towards} and ABINet~\cite{fang2021read}, training is divided into multi-stages. In our re-implementation, we train them in a single stage. In detail, all models are trained for 300K iterations with a total batch size of 256 with exactly the same training data (as mentioned in Section 4.1) for fair comparison. 
It is a common practice for Vision Transformers~\cite{DBLP:conf/iclr/DosovitskiyB0WZ21} to normalize the input at the beginning of each block. Following~\cite{xiong2020layer}, we re-implemented the ABINet-SV~\cite{fang2021read} with pre-normalization configuration (\ie, locates the layer normalization inside the residual blocks).

\section{More Results}
\subsection{Ablation Study}
\begin{table}[t]
\centering
\begin{tabular}{c|c|c|c|c}
\multicolumn{5}{c}{\includegraphics[width=0.16\linewidth]{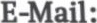}}  \\  
\multicolumn{5}{c}{(a) Example figure with text ``E-Mail:''.}\\\hline
Character   set & D+L & D+L+P & D+L+U &D+L+U+P      \\ \hline
Label & email & e-mail: & EMail & E-Mail: \\\hline
\multicolumn{5}{c}{(b) Labels under different character set.}\\
\end{tabular}
\caption{Example results of label processing. \textit{D}, \textit{L}, \textit{U} and \textit{P} denotes digits, lowercase, uppercase and punctuation respectively. } 
\label{Fig:label_processing}
\end{table}

\textbf{The necessity to search for the optimal $\alpha_{real}$}.  One of major differences between our training method and the existing ones is that we try to optimize the proportion of training data sampled from real datasets (\ie, $\alpha_{real}$) by protocol searching. We conduct the following experiment to demonstrate it is necessary to search for the optimal $\alpha_{real}$. Specifically, we only change the value of $\alpha_{real}$ of our baseline training protocol and record the corresponding recognition accuracy.

\begin{figure}
	    \centering
        \begin{subfigure}[t]{0.6\linewidth}
			\includegraphics[width=1\linewidth]{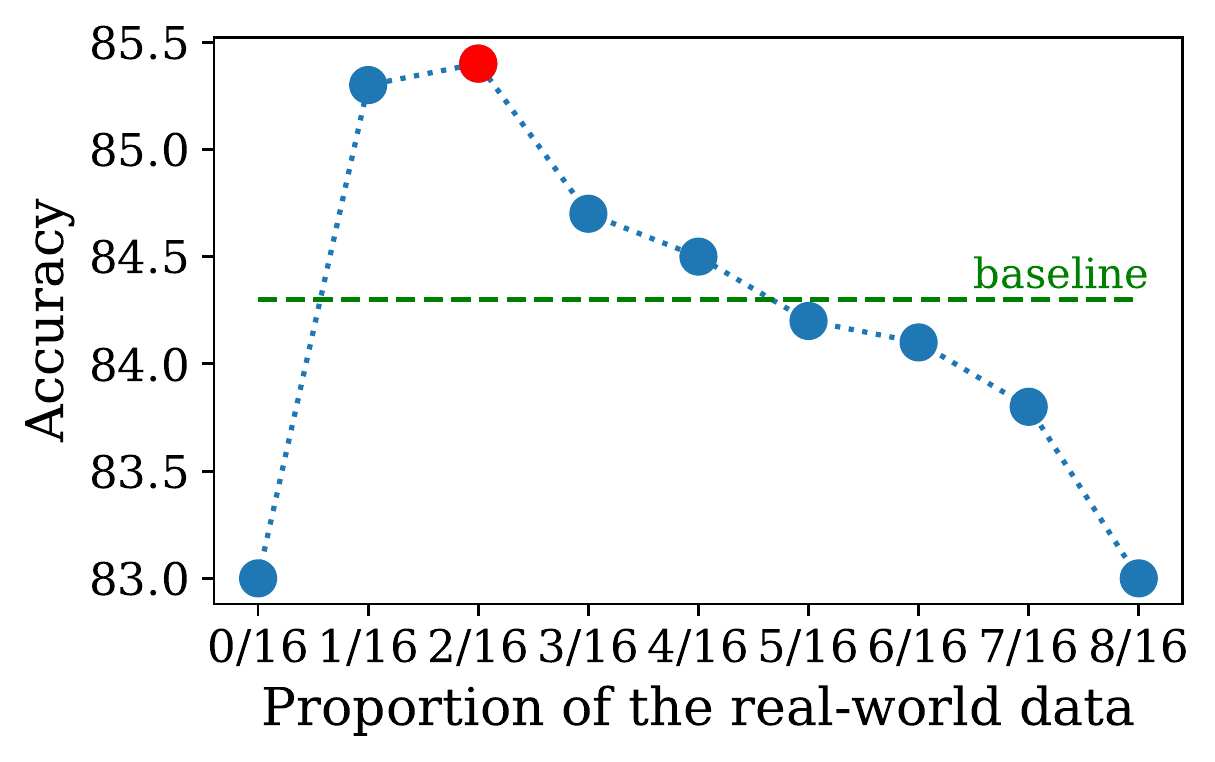}  
			\vskip -0.03in
			\caption{$\alpha_{real}$-accuracy}
		    \label{fig:real_ab}
		\end{subfigure}
        \hfill
        \begin{subfigure}[t]{0.39\linewidth}
			\includegraphics[width=1\linewidth]{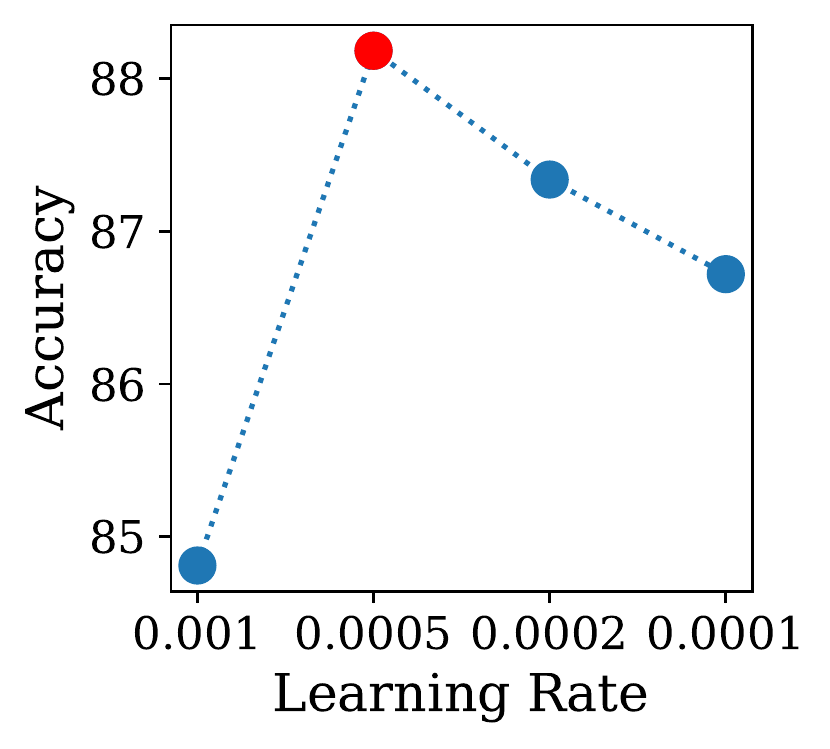}  
			\vskip -0.03in
			\caption{$lr$-accuracy}
			\label{fig:lr_ab}
		\end{subfigure}
	    \vskip -0.1in
        \caption{Ablation study for $\alpha_{real}$ and learning rate~($lr$). We mark the choice of our searched training protocol in red. (a)~Results of CRNN when training with different proportion $\alpha_{real}$ of the real-world data. (b)~Results of CRNN and Transformer baseline when training with different $lr$.} 
        \label{fig:extra_ab}
\end{figure}

Figure~\ref{fig:extra_ab}(a) illustrate how the recognition accuracy varies with $\alpha_{real}$: 1) when $\alpha_{real}$ increases from 0 to 0.5, the recognition accuracy first becomes better and then drops; 2) the optimal result is achieved when $\alpha_{real}=0.125$. These results first show that it is necessary to search for the optimal $\alpha_{real}$. 
Essentially, the distribution of the real datasets is closer to the real test images than the synthetic datasets, so using more real data for training can make the trained model perform better on the real test images. However, since the number of real training images is much smaller than the synthetic ones, using too higher ratio of real training images will lead to overfitting and finally harms the performance of the trained model. Besides, our search algorithm successfully finds the optimal $\alpha_{real}$, which demonstrates the effectiveness of our search algorithm once again.

\textbf{The effect of learning rate.}
To investigate the impact of different initial learning rate ($lr$) on the model performance, we test performance change of CRNN under different $lr$. In detail, we use the searched training protocols while only change $lr$ to different settings. As shown in Figure~\ref{fig:extra_ab}(b), when $lr$ decreases from 0.001 to 0.0001, the recognition accuracy first becomes better and then drops and the optimal result is achieved when ${lr}=0.0005$. 
Our search algorithm is able to find the optimal $lr$, demonstrating the effectiveness of the proposed search algorithm once again.

\subsection{Qualitative results}
Compared with baseline training protocol, our searched ones improves the recognition accuracy of various scene text recognition models. In Figure~\ref{fig:visual}, we visualize some results of our top-2 model (\ie, STAR-Net and TRBA-Net) under different training protocol.
Observing the results using our searched training protocol, we discover that hard examples with rare font styles and hazy appearance can also be accurately recognized, which indicates that optimizing training protocol is a potential route for accurate scene text recognition.

\begin{table}[t]
\addtolength{\tabcolsep}{3pt}
\centering
\begin{tabular}{c|cc|cc}
\toprule
\multirow{2}{*}{Images} & \multicolumn{2}{c|}{STAR-Net~\cite{liu2016star}} &  \multicolumn{2}{c}{TRBA-Net~\cite{baek2019wrong}}  \\
                        & Baseline               & Searched               & Baseline                      & Searched \\ \midrule
\multirow{3}{*}{\includegraphics[width=0.29\linewidth]{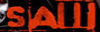}}                     & \multirow{3}{*}{sa\color{red}{l}}                      & \multirow{3}{*}{saw}                      & \multirow{3}{*}{sa\color{red}{ll}}                             & \multirow{3}{*}{saw}        \\ &&&&\\
&&&&\\ 
\multirow{3}{*}{
\includegraphics[width=0.29\linewidth]{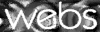} }                     & \multirow{3}{*}{w{\color{red}a}bs}                      & \multirow{3}{*}{webs}                      & \multirow{3}{*}{we{\color{red}ls}s}                             & \multirow{3}{*}{webs}        \\ 
&&&& \\
&&&& \\
\multirow{3}{*}{
\includegraphics[width=0.29\linewidth]{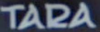} }                     & \multirow{3}{*}{ta{\color{red}2}a}                      & \multirow{3}{*}{tara}                      & \multirow{3}{*}{ta{\color{red}124}}                             & \multirow{3}{*}{tara}        \\
&&&&\\
&&&& \\
\multirow{3}{*}{
\includegraphics[width=0.29\linewidth]{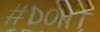} }                     & \multirow{3}{*}{{\color{red}h}do{\color{red}h}t}                      & \multirow{3}{*}{dont}                      & \multirow{3}{*}{do{\color{red}h}t}                              & \multirow{3}{*}{dont}        \\
&&&&\\
&&&& \\
\multirow{3}{*}{
\includegraphics[width=0.29\linewidth]{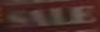} }                     & \multirow{3}{*}{{s\color{red}i}le}                      & \multirow{3}{*}{sale}                      & \multirow{3}{*}{{s\color{red}i}le}                               & \multirow{3}{*}{sale}        \\
&&&&\\
&&&& \\
\multirow{3}{*}{
\includegraphics[width=0.29\linewidth]{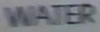} }                     & \multirow{3}{*}{{\color{red}m}ater}                      & \multirow{3}{*}{water}                      & \multirow{3}{*}{{\color{red}m}ater}                              & \multirow{3}{*}{water}        \\
&&&&\\
&&&& \\
\multirow{3}{*}{
\includegraphics[width=0.29\linewidth]{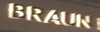} }                     & \multirow{3}{*}{braun{\color{red}i}}                      & \multirow{3}{*}{braun}                      & \multirow{3}{*}{braun{\color{red}i}}                             & \multirow{3}{*}{braun}        \\
&&&& \\
&&&& \\
\multirow{3}{*}{
\includegraphics[width=0.29\linewidth]{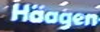} }                     & \multirow{3}{*}{h{\color{red}o}agen}                      & \multirow{3}{*}{haagen}                      & \multirow{3}{*}{ha{\color{red}g}gen}                             & \multirow{3}{*}{haagen}        \\
&&&& \\
&&&& \\
\multirow{3}{*}{
\includegraphics[width=0.29\linewidth]{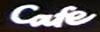} }                     & \multirow{3}{*}{ca{\color{red}d}e}                      & \multirow{3}{*}{cafe}                      & \multirow{3}{*}{ca{\color{red}u}e}                             & \multirow{3}{*}{cafe}        \\
&&&&\\
&&&& \\
\multirow{3}{*}{
\includegraphics[width=0.29\linewidth]{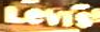} }                     & \multirow{3}{*}{{le\color{red}n}s}                      & \multirow{3}{*}{levis}                      & \multirow{3}{*}{{le\color{red}n}s}                             & \multirow{3}{*}{levis}        \\
&&&&\\
&&&& \\
\multirow{3}{*}{
\includegraphics[width=0.29\linewidth]{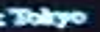} }                     & \multirow{3}{*}{{to\color{red}l}kyo}                      & \multirow{3}{*}{tokyo}                      & \multirow{3}{*}{{to\color{red}ir}yo}                             & \multirow{3}{*}{tokyo}        \\
&&&&\\
&&&& \\
\multirow{3}{*}{
\includegraphics[width=0.29\linewidth]{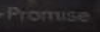} }                     & \multirow{3}{*}{{prom\color{red}u}se}                      & \multirow{3}{*}{promise}                      & \multirow{3}{*}{{prom\color{red}u}se}                             & \multirow{3}{*}{promise}        \\
&&&&\\
&&&& \\
\multirow{3}{*}{
\includegraphics[width=0.29\linewidth]{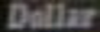} }                     & \multirow{3}{*}{{\color{red}la}llar}                      & \multirow{3}{*}{dollar}                      & \multirow{3}{*}{d{\color{red}a}llar}                             & \multirow{3}{*}{dollar}        \\
&&&&\\
&&&& \\
\multirow{3}{*}{
\includegraphics[width=0.29\linewidth]{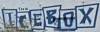} }                     & \multirow{3}{*}{i{\color{red}e}e{\color{red}e}x}                      & \multirow{3}{*}{icebox}                      & \multirow{3}{*}{{\color{red}tre}e{\color{red}d}ox}          & \multirow{3}{*}{icebox}        \\
&&&&\\
&&&& \\\bottomrule
\end{tabular}
\caption{Recognition results of some challenging examples. Notably, the models trained with our searched protocol can successfully handle these challenging examples.}
\label{fig:visual}
\end{table}

%% file: tables/evolution.tex
\algrenewcommand\algorithmicrequire{\textbf{Input:}}
\algrenewcommand\algorithmicensure{\textbf{Return:}}

\begin{algorithm}[t]
%   \scriptsize
\small
   \begin{algorithmic}[1]
      \Require hyperparameter configuration space $\Lambda$,  number of initial candidates $M_{init}$, population size $M$, parent population size $k$, max iteration $T$, Mutation probability $prob$, training dataset $\mathrm{D}_{train}$, validation dataset $\mathrm{D}_{val}$
    \State $\mathcal{P}\leftarrow \emptyset$    %\Comment{The population}
 \For{$i = 1,2,\ldots M_{init}$} % \Comment{Initialize population}
    \State $candidate \leftarrow \text{Random\_Sampling}(\Lambda)$
    \State $\text{Evaluate}(candidate, \mathrm{D}_{train}, \mathrm{D}_{val})$
    \State add $candidate$ to $\mathcal{P}$
 \EndFor
 \State $\mathcal{P}_{parent} \leftarrow  \text{Topk\_Selection}(\mathcal{P}, topk)$
 \For{$t = 1,2,\ldots T$}   \Comment{Evolve for $T$ iterations}
    \State $\mathcal{C} \leftarrow \emptyset$   \Comment{The childs}
     \For{$i = 1,2,\ldots M/2$} \Comment{Crossover}
        \State $candidate \leftarrow \text{Crossover}(\mathcal{P}_{parent})$
        \State $\text{Evaluate}(candidate, \mathrm{D}_{train}, \mathrm{D}_{val})$
        \State add $candidate$ to $\mathcal{P}$
     \EndFor
     \For{$i = 1,2,\ldots M/2$} \Comment{Mutation}
        \State $candidate \leftarrow \text{Mutation}(\mathcal{P}_{parent}, prob, \Lambda)$
        \State $\text{Evaluate}(candidate, \mathrm{D}_{train}, \mathrm{D}_{val})$
        \State add $candidate$ to $\mathcal{P}$
     \EndFor
     \State $\mathcal{P}_{parent} \leftarrow  \text{Topk\_Selection}(\mathcal{P}, topk)$
 \EndFor
 
      \Ensure the hyperparameter configuration with highest accuracy in $\mathcal{P}$
   \end{algorithmic}
    \caption{Evolutionary Hyperparameter Optimization}
\label{alg:evolution}
\end{algorithm}